\documentclass[a4paper, 10 pt, conference]{ieeeconf} 
\IEEEoverridecommandlockouts                              

\usepackage{graphicx}
\usepackage{hyperref}
\usepackage{acronym}
\usepackage{xcolor}
\usepackage{tikz}
\usepackage{siunitx}
\usepackage{booktabs}
\usepackage{pifont}
\newcommand{\cmark}{\ding{51}}%
\newcommand{\xmark}{\ding{55}}%
\usepackage{amsmath}
\usepackage[capitalize]{cleveref}
\usepackage{gensymb} 

\newacro{lidar}[LiDAR]{Light Detection and Ranging}
\newacro{ir}[IR]{Infrared}
\newacro{nir}[NIR]{Near Infrared}
\newacro{fov}[FoV]{Field of View}
\newacro{gpu}[GPU]{Graphics Processing Unit}
\newacro{tof}[ToF]{Time of Flight}
\newacro{cnn}[CNN]{Convolutional Neural Network}
\newacro{hri}[HRI]{Human Robot Interaction}
\newacro{nms}[NMS]{Non-maximum Suppression}

\usepackage[style=ieee,hyperref,natbib=true,backend=bibtex,firstinits,doi=false,%
     mincitenames=1,maxcitenames=2,maxbibnames=99,sorting=none,terseinits=false,hyperref=true]{biblatex}
\bibliography{references.bib}
\renewbibmacro*{bbx:savehash}{}
\defbibheading{bibliography}[\bibname]{\section*{References}}

\usepackage{makecell}
\usepackage{multirow}

\usetikzlibrary{positioning}

\title{\LARGE \bf
Person Segmentation and Action Classification for Multi-Channel Hemisphere Field of View LiDAR Sensors
}

\author{Svetlana Seliunina$^{*}$, Artem Otelepko$^{*}$, Raphael Memmesheimer, and Sven Behnke
\thanks{*Equal contribution}
\thanks{This work was funded by grant BE 2556/16-2 (Research Unit FOR 2535 Anticipating Human Behavior) of the German Research Foundation (DFG)}
\thanks{All authors are with the Autonomous Intelligent Systems group of University of Bonn, Germany; {\tt memmesheimer@ais.uni-bonn.de}}%
}

\begin{document}

\maketitle
\thispagestyle{empty}
\pagestyle{empty}

\begin{abstract}
Robots need to perceive persons in their surroundings for safety and to interact with them.
In this paper, we present a person segmentation and action classification approach that operates on 3D scans of hemisphere field of view LiDAR sensors. We recorded a data set with an Ouster OSDome-64 sensor consisting of scenes where persons perform three different actions and annotated it.
We propose a method based on a MaskDINO model to detect and segment persons and to recognize their actions from combined spherical projected multi-channel representations of the LiDAR data with an additional positional encoding.
Our approach demonstrates good performance for the person segmentation task and further performs well for the estimation of the person action states walking, waving, and sitting. An ablation study provides insights about the individual channel contributions for the person segmentation task.
The trained models, code and dataset are made publicly available.
\end{abstract}

\section{Introduction}

Perceiving persons in the environment is a crucial task for many applications, such as autonomous driving, service robots, and smart buildings.
\ac{lidar} sensors are promising for the detection and segmentation of persons in the surrounding for various reasons: ~i) \ac{lidar} measurements are more reliable than camera-based depth estimates. Hence, they allow for robust detection and precise localization. ~ii) As \ac{lidar} sensors are actively transmitting laser beams and interpreting their reflections, they remain largely unaffected by changes in lighting conditions and function in complete darkness. ~iii) \ac{lidar} sensors don't capture direct personal information like facial details, which can improve acceptance by the general public.

Person detection and segmentation are well studied for various sensor data modalities like RGB images~\cite{DBLP:journals/pami/RenHG017,DBLP:conf/cvpr/RedmonDGF16, shin:CVPR2024Wham}, RGB-D images~\cite{jafari2014real, xu2023onboard}, \ac{ir} images~\cite{DBLP:conf/eccv/DubailPMAGP22,DBLP:conf/fgr/ClapesJME20,zhang:PAMI2023visibleIR}, thermal images~\cite{WagnerFHB:ESANN2016,DBLP:conf/fgr/ClapesJME20,tian:TMM2024RGB-Thermal} and 2D-\ac{lidar} sensors~\cite{beyer2018deep, jia:IROS2020dr, yang2024li2former}. With the creation of larger datasets for semantic segmentation of 3D-\ac{lidar} sensors~\cite{DBLP:conf/iccv/BehleyGMQBSG19} methods for the semantic segmentation of relevant classes such as cars and persons became of increasing interest~\cite{yan2020online, LiuKCCZPCL23, hong:PAMI2024unified}. 
A point-level segmentation allows for precise person localization and further serves as input for  tracking~\cite{DBLP:journals/ras/WangWLMY17} and human pose estimation~\cite{DBLP:journals/cacm/ShottonSKFFBCM13}.
Especially in the context of assistive service robots, the understanding of the activities of the surrounding persons is of interest. 
While current \ac{lidar}-based approaches focus solely on the detection and segmentation, we aim to extend the understanding to an activity level. 
\begin{figure} 
    \centering
    \footnotesize
    \includegraphics[width=\linewidth]{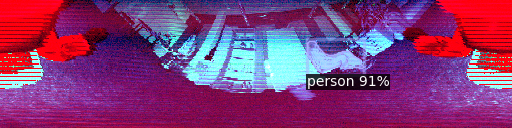}
    Person segmentation\\
    \vspace{0.15cm}
    \includegraphics[width=\linewidth]
    {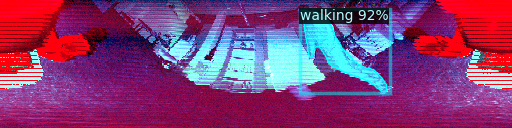}
    Action classification
    \vspace{-0.2cm}
    \caption{We segment persons and classify their actions from spherical projected 2D representations of multi-channel hemisphere \ac{fov} \ac{lidar}.}
    \label{fig:teaser}
    \vspace{-.5cm}
\end{figure}

Recent research further has demonstrated that models trained in a supervised fashion could potentially leak private information from training data~\cite{DBLP:conf/ccs/HitajAP17,li2023multi}. We argue that when developing approaches based on sensors that do not measure high-resolution images in the first place, it is impossible to leak such private information.

Hemisphere \ac{lidar} sensors like the Ouster OSDome cover 180\degree~\ac{fov} with 64 or 128 laser beams with range, signal, reflectivity and \ac{nir} channels.
As shown in Fig.~\ref{fig:teaser}, its measurements can be represented in sensor coordinates as 2D matrix with rotation angle and beam number as coordinate axes.
The measurement directions are not evenly distributed across the hemispheric \ac{fov}, but are closer together near the axis of rotation and further apart at the periphery, which may pose a challenge for person detection.

In this paper, we explore if recent 2D image detection and segmentation models are capable of dealing with these unevenly distributed beams for person segmentation and action classification tasks.

\noindent The contributions of the paper are as follows:
\begin{itemize}
  \item We provide an annotated dataset for person segmentation and action classification acquired from a hemisphere field of view \ac{lidar} sensor.
    \item We present a person segmentation approach operating on the combined channel representations of the \ac{lidar} data and further extend this approach to estimate action states of a person, such as walking, waving, and sitting.
    \item The model, code and dataset are made publicly available to the community on Github\footnote{\url{https://github.com/AIS-Bonn/lidar_person_action_detection}}.
\end{itemize}

\section{Related Work}

Detecting persons from sensor data streams is a well-established research topic.
In the following, we review the state of the art in person detection from various sensor data modalities and put special emphasis on privacy-preserving person detection methods.
Person detection and segmentation methods have a strong focus on image-based methods. 
Historically, persons were detected with Haar feature cascade detectors~\cite{DBLP:conf/cvpr/ViolaJ01} and histogram of gradient~\cite{DBLP:conf/cvpr/DalalT05} methods. 
With increasing classification performance of 2D-\ac{cnn} they have soon been extended to detection~\cite{DBLP:journals/pami/RenHG017,DBLP:conf/cvpr/RedmonDGF16} and segmentation models \cite{cortinhal2020salsanext,DBLP:conf/iccv/BehleyGMQBSG19}. 


Sensors for privacy-preserving person detection range from \ac{lidar} to thermal cameras to specially developed sensors. The advances from image-based person detection have always been adapted to other sensors like 2D-\ac{lidar}~\cite{beyer2018deep} and 3D-\ac{lidar}~\cite{DBLP:conf/ivs/KidonoMWNM11}. 
Günter et al.~\cite{DBLP:conf/intenv/GunterBK020} present a privacy-preserving person detection approach for solid state \ac{lidar} sensors.
Dubail et al.~\cite{DBLP:conf/eccv/DubailPMAGP22} proposed privacy-preserving person detection using ultra-low resolution \ac{ir} cameras.
The ChaLearn Looking At People Challenge~\cite{DBLP:conf/fgr/ClapesJME20} focuses on depth gathered from an RGB-D camera and thermal images for the benchmarking of identity preserving approaches. The annotations are on a bounding box level, whereas we provide annotations on a pixel level. In the related challenge, the best performing approaches utilized the thermal images with a combination of Faster R-CNN~\cite{DBLP:journals/pami/RenHG017} + Faster R-DCN (ResNet-50)~\cite{DBLP:conf/iccv/DaiQXLZHW17} and a soft \ac{nms}. 


In many instances, especially for \ac{hri}, one might be interested not only in the persons' location but also in their gestures or activities.  
\textcite{DBLP:conf/icra/DroeschelSHB11} track persons of interest with a 2D \ac{lidar} and utilize a \ac{tof} camera to extract pointing gestures from a mobile manipulating platform. Their approach, due to the usage of \ac{tof} is not usable to preserve privacy, as the sensor leaks facial attributes. For gesture recognition, they segment the body into multiple parts and then estimate the face centroid, elbow position and hand position. Vectors between the body keypoints are utilized to estimate a showing and pointing gesture.

For recognizing activities directly from videos, spatio-temporal models have been proposed. 
SlowFast~\cite{DBLP:conf/iccv/Feichtenhofer0M19} proposes a two stream approach, one stream (high frame rate) focuses on extracting temporal features and the second stream (low frame rate) focuses on the extraction of detailed spatial semantics.
Multi-modal approaches that generalize well across different sensor modalities have been proposed for the action recognition task in a supervised setting~\cite{DBLP:conf/iros/MemmesheimerTP20} and for one-shot inference following a semi-supervised setting~\cite{DBLP:conf/icpr/MemmesheimerTP20}.
In contrast to the above-mentioned approaches, which focus on sequence classification, our proposed approach can also localize the activities.

3D \ac{lidar} semantic segmentation datasets such as SemanticKITTI~\cite{DBLP:conf/iccv/BehleyGMQBSG19} often focuses on automotive scenes. In contrast, our focus is on indoor person segmentation and additional action classification. 

\section{Dataset}

\begin{figure}[t]
  \centering
  \footnotesize
  \begin{tikzpicture}
      \node[anchor=south west,inner sep=0] (tiago) at (0,0) {\includegraphics[width=3.3cm]{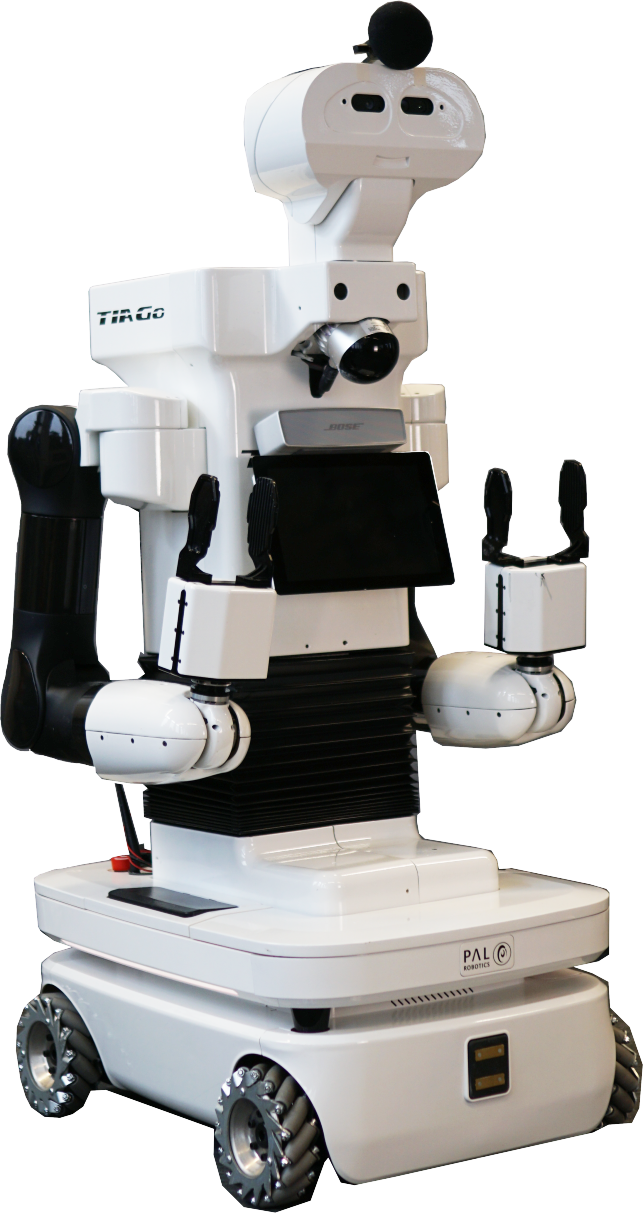}};
      \node (ouster) at (5,4.5) {\includegraphics[width=2cm]{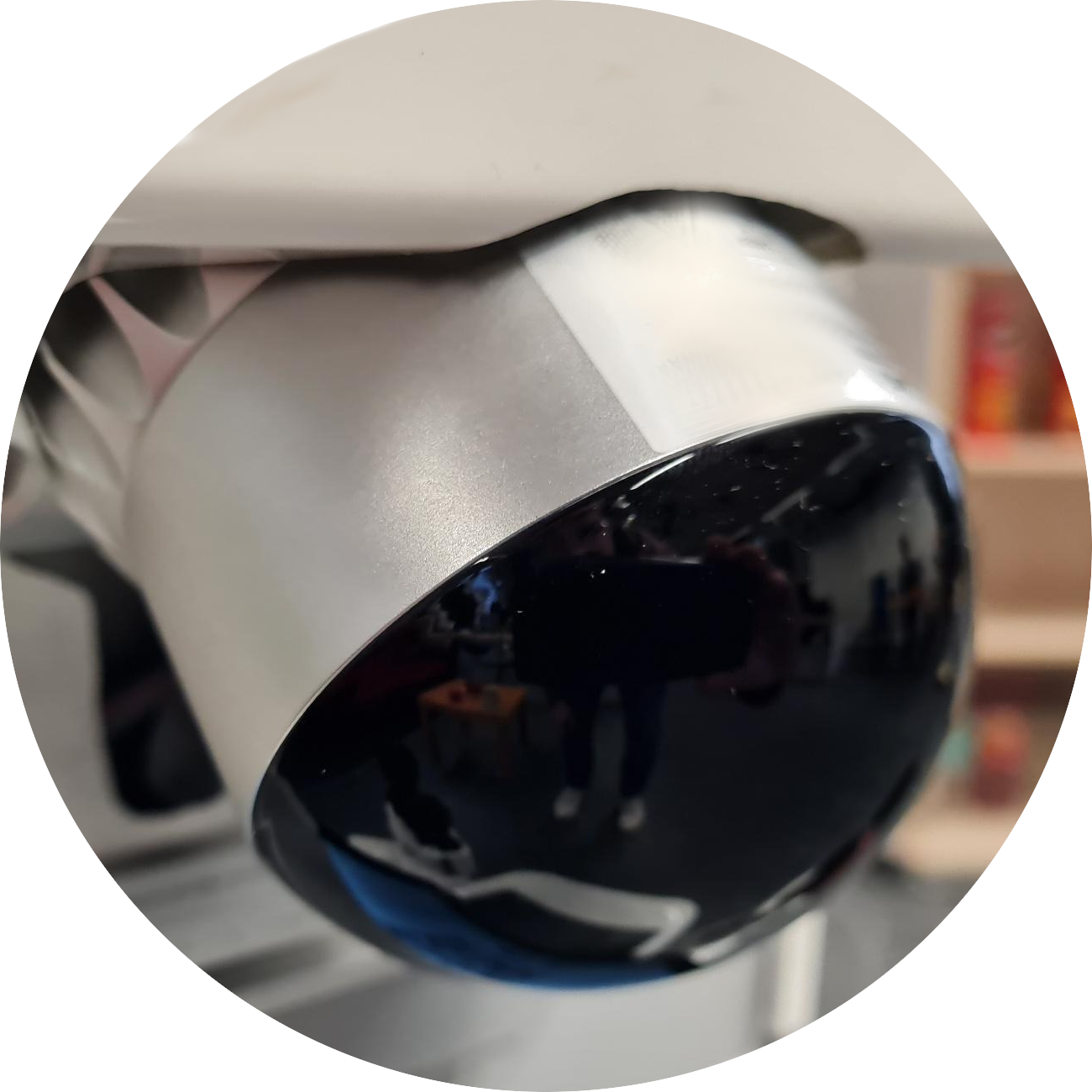}};
      \draw[->, thick] (ouster) -- (2.2,4.5);
      \node[below right=-.5cm and 0cm of tiago, node distance=0.1] (tiagotxt)  {PAL TIAGo++ Omni};
      \node[below] at (ouster.south) {Ouster OSDome-64};
    \end{tikzpicture}
  \vspace*{-2mm}
  \caption{The dataset collection robot setup.}
  \label{fig:dataset_setup}
\end{figure}
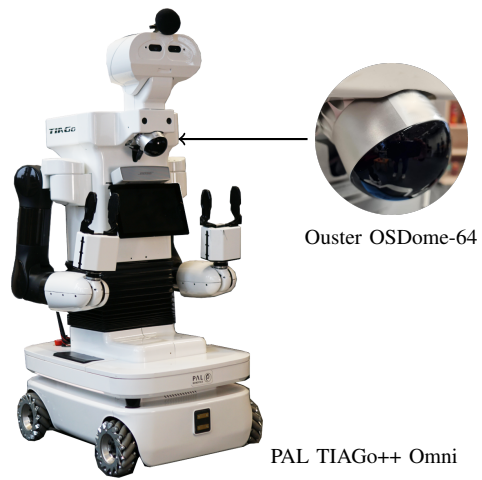

\begin{figure}[t]
  \centering
  \footnotesize
  \includegraphics[width=\linewidth]{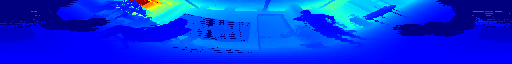}\\
   \vspace*{-.5mm}
   Range\\
  \vspace{0.1cm}
  \includegraphics[width=\linewidth]{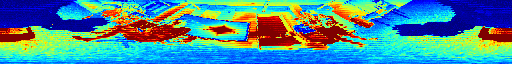}\\
   \vspace*{-.5mm}
  Signal\\
  \vspace{0.1cm}
  \includegraphics[width=\linewidth]{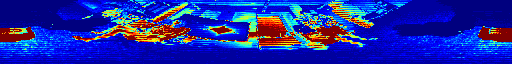}\\
   \vspace*{-.5mm}
  Reflectivity\\
  \vspace{0.1cm}
  \includegraphics[width=\linewidth]{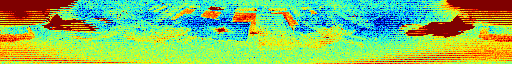}\\
   \vspace*{-.5mm}
  \ac{nir}\\
  \vspace{-0.2cm}
  \caption{Example data from the individual Ouster OSDome-64 channels.}
  \label{fig:sensor_channels}
\end{figure}

\begin{figure}[t]
  \centering
  \includegraphics[width=\linewidth]{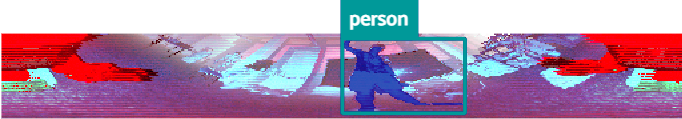}\\
  \vspace{0.2cm}
  \includegraphics[width=\linewidth]{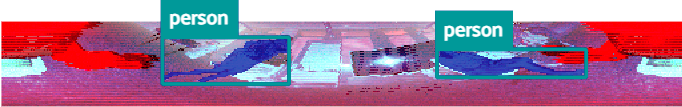}\\
  \vspace{0.2cm}
  \includegraphics[width=\linewidth]{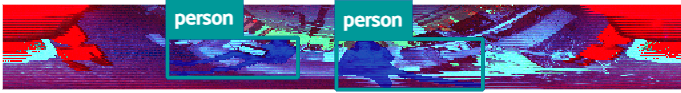}\vspace*{-1ex}
  \caption{Person segmentation examples with ground truth masks (blue).}
  \label{fig:dataset_detection}
\end{figure}

\begin{figure}[t]
  \centering
  \includegraphics[width=\linewidth]{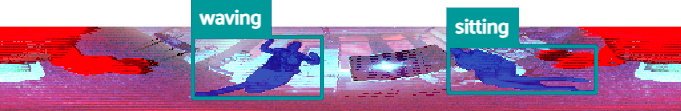}\\
  \vspace{0.1cm}
  \includegraphics[width=\linewidth]{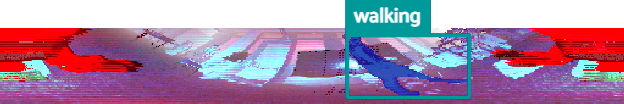}\\
  \vspace{0.1cm}
  \includegraphics[width=\linewidth]{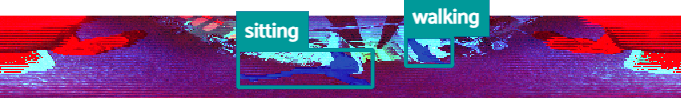}\vspace*{-1ex}
  \caption{Action recognition examples with ground truth mask (blue).}
  \label{fig:dataset_actions}
  \vspace*{-1.5ex}
\end{figure}

The dataset is collected with an Ouster OSDome-64 \ac{lidar} attached on a TIAGo++ omnidirectional mobile robot.
The \ac{lidar} sensor is attached in the front and tilted slightly downwards to capture the environment in front of the robot, as depicted in Fig.~\ref{fig:dataset_setup}. 

The Ouster OSDome-64 \ac{lidar} provides four channels, shown in \cref{fig:sensor_channels}: 1) the distance of the measured surface in mm, 2) the signal intensity (number of photons in the signal return measurement) 3) reflectivity (scaled intensity  
based on measured range and sensor sensitivity at that range) 4) \ac{nir} (photons related to natural environmental illumination). 

The dataset contains 442 scans of persons in different action states, such as walking, waving, and sitting. The dataset samples are randomly divided once into training, validation and test datasets with proportions of 70/15/15.
The dataset consists of scenes of different complexity. In some of them, only one or two persons are present, and they are clearly visible. In others, more people are present in the background. Persons of different body types and sizes with different clothes are present in the data. The background itself changes from scene to scene, and people are often partially occluded by various objects. It is also worth mentioning that the data is collected with both the stationary and the moving robot.

During the walking action, people walked in different directions but were instructed not to raise their hands. While waving, the subjects could stand or walk, but one or both hands were always raised above waist height and moving. Finally, for the sitting action, people could sit in different positions, be occluded by a table and move the chair. In each action sequence, sensor data was collected ten times.

To support the labeling, we used a semi-automatic labeling approach based on SegmentAnything~\cite{DBLP:conf/iccv/KirillovMRMRGXW23} and manually corrected the masks.
Examples for person segmentation and action classification are shown in Figs.~\ref{fig:dataset_detection} and \ref{fig:dataset_actions}, respectively.

\section{Method}

We adapted a MaskDINO model~\cite{DBLP:conf/cvpr/Li0XL0NS23} to jointly train a person detection and segmentation model on our proposed multi-channel representation. 

\subsection{Representation}

Each measurement channel has advantages and disadvantages in terms of perceptual separation of individuals, depending on the situation. None of the channels consistently provides universal perceptual separation for individuals, though. Hence, we incorporated all  \ac{lidar} channels in the person perception, thereby improving the separation of individuals, as the quality of the masks was significantly influenced by this. 

After conducting several experiments with various combinations of channels, we decided to utilize reversed range data as the opacity channel for visualization. This technique enabled us to render objects that are located far from the sensor transparent in tools that support four channel images, and to achieve subjectively good perceptual separation of the individuals in the image in tools that do not correctly support four channel images. This was advantageous mainly for simplifying the process of manual pixel-level annotation.

The resulting image representation consists of \ac{nir}, reflectivity, signal and the reversed range in separate channels resulting in a four channel image of size 512$\times$64.

\begin{figure}[t]
    \centering
    \includegraphics[width=.95\linewidth]{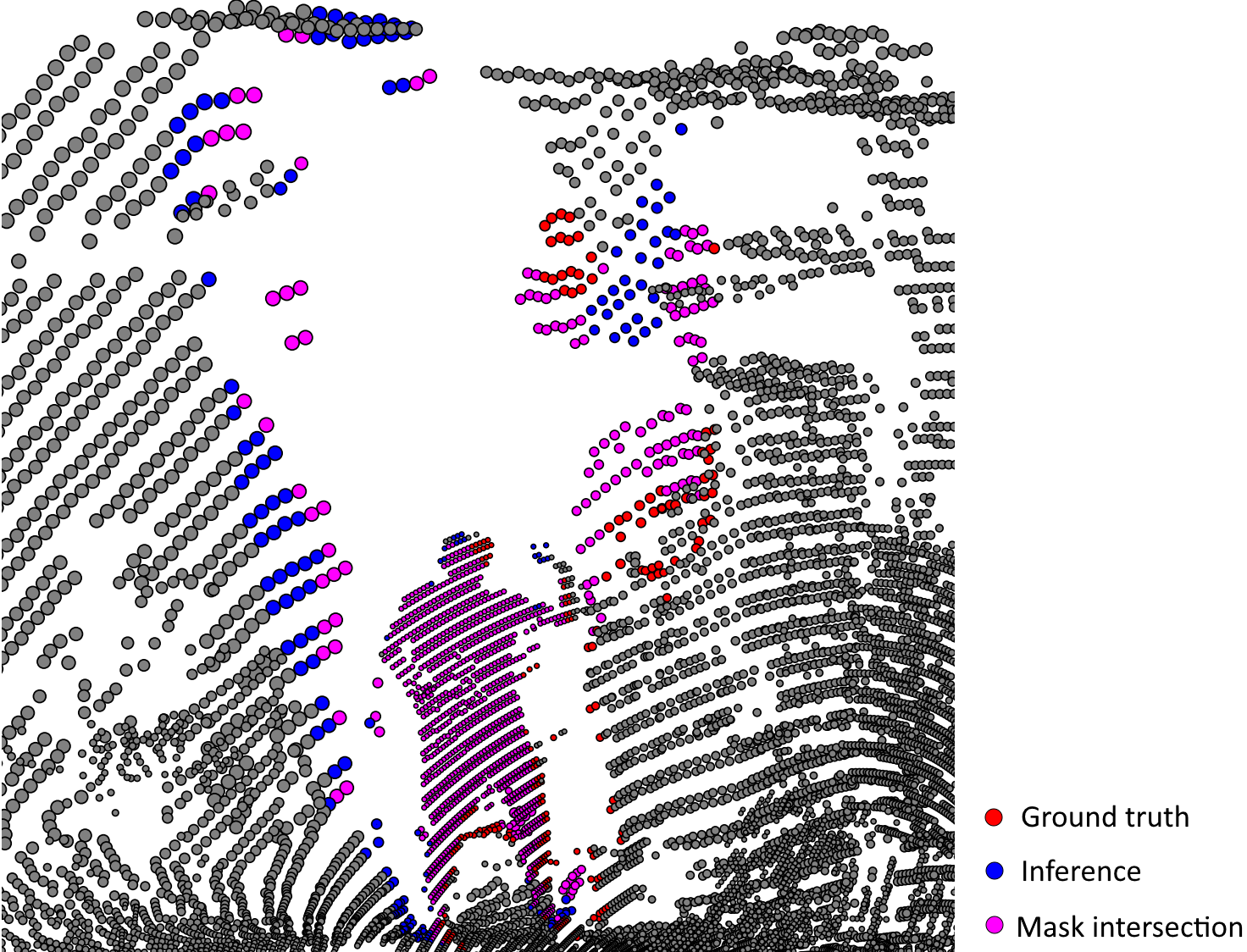}
    \vspace*{-1.5ex}
    \caption{Point cloud with ground truth and segmentation masks.}
    \label{fig:point_cloud}
    \vspace*{-1.5ex}
\end{figure}

From the range image and \ac{lidar} metadata, we reconstructed the Euclidean coordinates of each corresponding pixel. The sample image together with the ground truth and inferred masks are presented in \cref{fig:point_cloud}. Errors in the ground truth mask are the consequences of using the semiautomatic approach for labeling on the four-channel image. 

Positional encoding of the XYZ point coordinates were added as additional input channels, resulting in a seven-channel input image. 
The positional encoding is computed using Ouster Sensor SDK as follows:
\begin{gather*}
    \begin{aligned}
    x &= (r - |n|) \cos(\theta_{enc} + \theta_{azi}) \cos(\phi) + x_n \cos(\theta_{enc}), \\
    y &= (r - |n|) \sin(\theta_{enc} + \theta_{azi}) \sin(\phi) + x_n \cos(\theta_{enc}), \\
    z &= (r - |n|) \sin(\phi) + z_n,
    \end{aligned}
\end{gather*}
with  \vspace*{-2ex}   
\begin{gather*}
    |n| = \sqrt{x_n^2 + z_n^2}, \quad
    \theta_{enc} = 2 \pi \left(1 - \frac{i}{w}\right), \\
    \theta_{azi} = - 2 \pi \frac{\alpha_i}{360}, \quad
    \phi = 2 \pi \frac{\beta_i}{360},
\end{gather*}
where $r$ is the range value of the measurement ID $i$. Parameters $x_n$,  $z_n$ describe the distance from the center of the \ac{lidar} origin coordinate frame to its front optics.  $w$ denotes the scan width. $\alpha_i$, $\beta_i$ denote the azimuth angle and altitude angle of beam $i$, respectively.

\subsection{Model}

We adapted MaskDINO~\cite{DBLP:conf/cvpr/Li0XL0NS23} with a SwinL transformer~\cite{DBLP:conf/iccv/LiuL00W0LG21} backbone. 
The model operates on four-channel input images, or seven channels if the positional encoding is used, by employing a convolutional encoding layer before the backbone in both the trainer and the predictor. The full representations are then fed to the model during training and inference. The same model is applicable for person segmentation and action detection.
The hyperparameters in our approach were based on the MaskDINO configuration with additional changes to incorporate the images with required number of channels and resizing to 512$\times$64 pixels. Horizontal random flip was used as data augmentation. The models were trained on a Nvidia RTX 3090 GPU, which resulted in approximately 35 minutes training time for 5,000 iterations.

\begin{table}
\renewcommand{\arraystretch}{1.3}
\caption{Training results for person detection}
\vspace*{-1ex}
\label{table:person_detection}
\centering

\begin{tabular}{c|r|r|r|r}
    \toprule
    \multirow{2}{*}{\makecell{\bf{Pos.}\\ \bf{enc.}$^*$}} & \bf{Precision} & \bf{F1-score} & \bf{Precision} & \bf{F1-score}\\
    & \multicolumn{2}{c|}{Frozen Backbone} & \multicolumn{2}{c}{Backbone also trained} \\
    \midrule
    \xmark & 0.98 & 0.93 & 0.98 & 0.97\\
    \cmark & 0.98 & 0.95 & 0.98 & 0.97\\
    \bottomrule
\end{tabular}\\
\vspace{0.1cm}
$^*$positional encoding
\vspace*{-0.5cm}
\end{table}

\section{Evaluation}

We evaluate our approach for person segmentation and action classification on the proposed dataset. We further give an ablation study on the channel contributions and the effect of the positional encoding for the person segmentation task. Finally, we analyze the applicability on an actual robot setup for online person segmentation and action classification.

\subsection{Person Segmentation}

The first set of experiments performs person segmentation with all four channels from the constructed dataset. Our training procedure initializes MaskDINO with SwinL transformer backbone with weights pretrained on COCO 2017.

The models were trained several times for 5,000 iterations with a step at 4,000 iterations, which multiplied the learning rate by 0.1. 
We trained the models with base learning rates \num{1e-3}, \num{1e-4}, and \num{1e-5}
and found that models trained with a \num{1e-4} learning rate provided the most promising and consistent results. After determining the base learning rate, we conducted several experiments with a frozen and unfrozen backbone. 

Using a frozen backbone provides good results, but the F1-score varies between 0.6 and 0.9. To further improve the results, we decided to unfreeze the backbone.
A backbone multiplier of \num{1e-5} provided the best results with an F1-score higher than 0.9.

The implementation of positional encoding allowed us to improve the performance of the trained models, resulting in higher precision and F1-score.

The best attempts to train the model for person segmentation with and without positional encoding are listed in \cref{table:person_detection}. These results correspond to the final iteration of training for 5,000 iterations with the base learning rate of \num{1e-4}.

\begin{figure}[t]
  \centering
  \includegraphics[width=\linewidth]{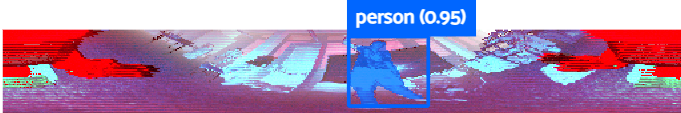}\\
  \vspace{0.2cm}
  \includegraphics[width=\linewidth]{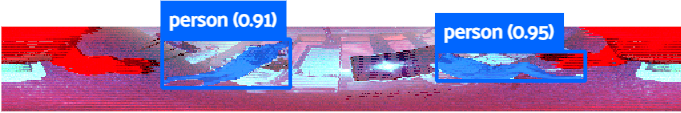}\\
  \vspace{0.2cm}
  \includegraphics[width=\linewidth]{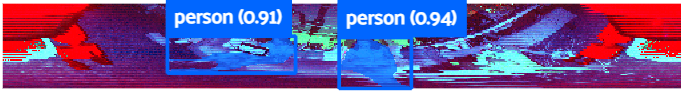}\vspace*{-1ex}
  \caption{Person detection examples with inferred mask (blue).}
  \label{fig:detections_person}
  \vspace{-0.1cm}
\end{figure}

The models trained with the base learning rate equal to \num{1e-4} and unfrozen backbone with the backbone multiplier equal to \num{1e-5} provided good results and were used in the following experiments. \cref{fig:detections_person} depicts person segmentation on the sample images from \cref{fig:dataset_detection} using the model without positional encoding.

\begin{figure}[t]
  \centering
  \includegraphics[width=\linewidth]{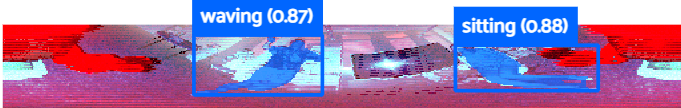}\\
  \vspace{0.2cm}
  \includegraphics[width=\linewidth]{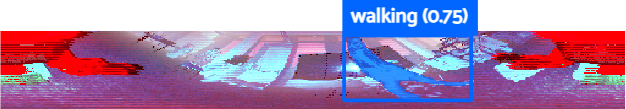}\\
  \vspace{0.2cm}
  \includegraphics[width=\linewidth]{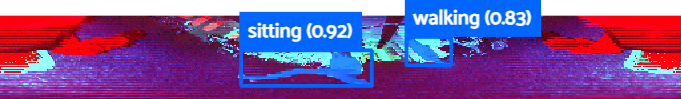}\vspace*{-1ex}
  \caption{Action detection examples inferred mask (blue).}
  \label{fig:detections_actions}
\end{figure}

\begin{table}[!t]
\renewcommand{\arraystretch}{1.3}
\caption{Action detection initialized with person detection weights.}
\vspace*{-1ex}
\label{table:actions_from_1_class}
\centering
\begin{tabular}{l|c|r|r|r|r}
    \toprule
    
    \multirow{2}{*}{\bf{Class}} & \multirow{2}{*}{\makecell{\bf{Pos.}\\ \bf{enc.$^*$}}}  & \bf{Precision} & \bf{F1-score} & \bf{Precision} & \bf{F1-score}\\
    & & \multicolumn{2}{c|}{Frozen Backbone} & \multicolumn{2}{c}{Non Frozen Backbone} \\
    
    \midrule
    sitting & \xmark & 0.97 & 0.95 & 1.00 & 0.97\\
    walking & \xmark &0.88 & 0.87 & 0.85 & 0.85\\
    waving & \xmark &0.83 & 0.89 & 0.86 & 0.91\\
    \cline{1-6}
    w. avg$^{**}$ & & 0.91 & 0.91 & 0.92 & 0.92\\
    \midrule
    sitting &\cmark & 0.95 & 0.95 & 0.95 & 0.96\\
    walking & \cmark &0.89 & 0.91 & 0.93 & 0.95\\
    waving & \cmark &0.92 & 0.92 & 0.86 & 0.91\\
    \cline{1-6}
    w. avg$^{**}$ & & 0.93 & 0.93 & 0.92 & 0.94\\
    \bottomrule
\end{tabular}\\
\vspace{0.1cm}\scriptsize
$^*$positional encoding, $^{**}$weighted average
\vspace{-0.5cm}
\end{table}

\subsection{Action Classification}

For domestic service robot applications, the action classification of the surrounding persons might be of interest. 
We initialized the weights from our person segmentation models and fine-tuned them to the action classification task.

Similarly to the person segmentation experiment, the models were trained several times for 5,000 iterations with a step at 4,000 iterations, which multiplied the learning rate by 0.1. 

The base learning rate in the experiments was chosen to be \num{1e-4} and \num{1e-5}. The model with learning rate \num{1e-5} performed slightly worse with an F1-score lower than 0.9, so we set the learning rate to \num{1e-4}.
After determining the base learning rate, we conducted several experiments with a frozen and unfrozen backbone. The backbone learning rate multiplier for the unfrozen model was set to \num{1e-5}.

The best attempts to train the model for action classification are listed in the \cref{table:actions_from_1_class}. These results correspond to the final iteration of training for 5,000 iterations with the base learning rate of \num{1e-4}.
Frozen and unfrozen backbones performed equally in this experiment. The positional encoding had a positive influence. \cref{fig:detections_actions} depicts action classification on the sample images from \cref{fig:dataset_actions} using the best model without positional encoding.

Another approach was to initialize the training with weights of a pre-trained on the COCO 2017 dataset MaskDINO model with SwinL transformer backbone instead of using the weights of a trained model for the person detection task.
To compare the results of different approaches fairly, we trained the models for 10,000 iterations with a base learning rate \num{1e-4} and a step at 8,000 iterations, which multiplied the learning rate by 0.1. The best attempts are shown in \cref{table:actions_from_maskdino}.

\begin{table}[!t]
\renewcommand{\arraystretch}{1.3}
\caption{Action detection models initialized with MaskDINO weights}
\vspace*{-1ex}
\label{table:actions_from_maskdino}
\centering
\begin{tabular}{l|c|r|r|r|r}
    \toprule
    \multirow{2}{*}{\bf{Class}} & \multirow{2}{*}{\makecell{\bf{Pos.}\\ \bf{enc.}$^*$}} & \bf{Precision}  & \bf{F1-score} & \bf{Precision} & \bf{F1-score}\\
    & & \multicolumn{2}{c|}{Frozen Backbone} & \multicolumn{2}{c}{Non Frozen Backbone} \\
    \midrule
    sitting & \xmark & 0.97 & 0.94 & 1.00 & 0.82\\
    walking & \xmark & 0.96 & 0.94 & 1.00 & 0.77\\
    waving & \xmark & 0.89 & 0.91 & 0.73 & 0.73\\
    \cline{1-6}
    w. avg$^{**}$ & & 0.95 & 0.93 & 0.92 & 0.78\\
    \midrule
    sitting &\cmark & 0.93 & 0.95 & 1.00 & 0.97\\
    walking & \cmark &1.00 & 0.94 & 0.88 & 0.82\\
    waving & \cmark &0.83 & 0.87 & 0.93 & 0.95\\
    \cline{1-6}
    w. avg$^{**}$ & & 0.92 & 0.93 & 0.94 & 0.92\\
    \bottomrule
\end{tabular}\\
\vspace{0.1cm}\scriptsize
$^*$positional encoding, $^{**}$weighted average\vspace*{-4ex}
\end{table}

Contrary to training the model using the weights of a trained model for the person detection task, this experiment provided the best results with frozen backbone. The results of this experiment, however, were significantly less consistent than the previous one.  

\subsection{Channel Contribution Ablation}

\begin{table}[!t]
\renewcommand{\arraystretch}{1.3}
\caption{Ablation on channel contribution (person detection task)}
\vspace*{-1ex}
\label{table:person_detection_channel_ablation}
\centering

\begin{tabular}{l|c|r|r|r r|r|r}
    \toprule
        \multirow{2}{*}{\makecell{\bfseries Excluded \\ \bfseries channel}} & \multirow{2}{*}{\makecell{\bf{Pos.}\\ \bf{enc.}$^*$}} & 
        \multicolumn{3}{c}{\bfseries Result}\\
        \cline{3-5}
        & & \bfseries Precision & \bfseries Recall & \bfseries F1-score \\
        
    \midrule
        - & \xmark & 0.89 & 0.42 & 0.57\\
        \cline{1-5}
        \ac{nir} & \xmark & 0.97 & 0.94 & 0.95\\
        Reflectivity & \xmark & 0.97 & 0.74 & 0.84\\
        Signal & \xmark & 0.96 & 0.78 & 0.86\\
        Range & \xmark & 0.65 & 0.67 & 0.66 \\
        
    \midrule
        - & \cmark & 0.83 & 0.63 & 0.72\\
        \cline{1-5}
        \ac{nir} & \cmark & 0.94 & 0.85 & 0.89\\
        Reflectivity& \cmark & 0.60 & 0.31 & 0.41\\
        Signal& \cmark & 0.93 & 0.85 & 0.89\\
        Range& \cmark & 0.58 & 0.33 & 0.42 \\
    \bottomrule
\end{tabular}\\
\vspace{0.1cm} \scriptsize
$^*$positional encoding
\end{table}

To understand the channel contributions of the proposed representation, we trained the models with a random fixed seed excluding different image channels and measured the performance of the model at the end of the training.
We conducted the ablation on the person segmentation task by utilizing the hyperparameters to match the best experiment from \cref{table:person_detection}. The results for each excluded channel are shown in \cref{table:person_detection_channel_ablation}.
\ac{nir} and signal channels have only a minor contribution while reflectivity and range channels data have higher contribution. Our ablation study indicates that the contribution to the performance of the model of different channels differs between the cases with positional encoding and without it. 
The ablation study shows that the model with fewer channels produces a better result than the one with all of them for both models. One possible explanation is that the one-layer convolutional encoder does not deal well with transforming our multi-channel input to the desired MaskDINO input format. In future work, we may explore different encoders and their contribution.

\begin{figure}[t]
    \footnotesize
  \centering
  \includegraphics[width=\linewidth]{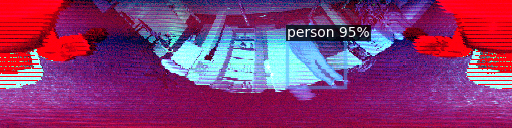}\\
  Person segmentation\\
  \vspace{0.1cm}
  \includegraphics[width=\linewidth]{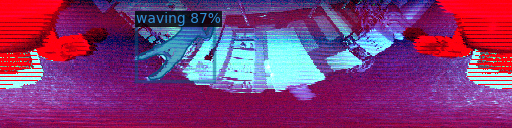}\\
   Action classification\\
   \vspace{-.2cm}
  \caption{Online detection experiment on the TIAGO++ Omni robot platform.}
  \label{fig:tiago_experiments}
  \vspace{-0.5cm}
\end{figure}
\subsection{Online Application on Robot}
To assess our methods for real-time application, we tested the inference of the best models for person segmentation (\cref{table:person_detection}) and action classification (\cref{table:actions_from_1_class}) on the data received from an Ouster OSDome-128 \ac{lidar} which yields images of size 512$\times$128. Note, the \ac{lidar} sensor for online experiments is slightly different and has the double amount of laser beams, demonstrating that our proposed approach generalizes across different versions of the sensor without adaptations. The average inference rate on a Zotac ZBOX QTG7A4500 equipped with an Nvidia RTX A4500 16GB GPU yields 11\,Hz with and without positional encoding, which is suitable e.g. as input for person tracking approaches. 
Examples of the results for person segmentation and action classification are shown in \cref{fig:tiago_experiments}. 
The results indicate that our approach generalizes well for images of different sizes from different sensors and is suitable for online application.

\section{Conclusion}

We presented a person segmentation and action classification approach for multi-channel data of hemisphere \ac{fov} 3D \ac{lidar} sensors.
A dataset with segmentation-level annotations on multi-channel \ac{lidar} measurements has been collected and annotated and a model based on MaskDINO has been adapted and trained to estimate person segments and has further been shown to be capable of estimating three different action classes relevant for \ac{hri}.
An ablation study provided insights into the channel contributions for the person segmentation model, demonstrating that the range and reflectivity channels as well as the positional encoding contribute significantly to the performance.
Our approach demonstrated good performance for both, the person segmentation and the classification of three different person states.
It is real-time capable and applicable to a sensor with more LiDAR beams.

\printbibliography

@inproceedings{DBLP:conf/eccv/DubailPMAGP22,
  author       = {Thomas Dubail and
                  Fidel Alejandro {Guerrero Pe{\~{n}}a} and
                  Heitor Rapela Medeiros and
                  Masih Aminbeidokhti and
                  Eric Granger and
                  Marco Pedersoli},
  title        = {Privacy-Preserving Person Detection Using Low-Resolution Infrared
                  Cameras},
  booktitle    = {European Conference on Computer Vision (ECCV) Workshops},
  series       = {Lecture Notes in Computer Science},
  volume       = {13805},
  pages        = {689--702},
  publisher    = {Springer},
  year         = {2022},
}

@inproceedings{shin:CVPR2024Wham,
  title={{WHAM}: Reconstructing world-grounded humans with accurate {3D} motion},
  author={Shin, Soyong and Kim, Juyong and Halilaj, Eni and Black, Michael J},
  booktitle={IEEE/CVF Conference on Computer Vision and Pattern Recognition (CVPR)},
  pages={2070--2080},
  year={2024}
}

@inproceedings{DBLP:conf/intenv/GunterBK020,
  author       = {Andrei G{\"{u}}nter and
                  Stephan B{\"{o}}ker and
                  Matthias K{\"{o}}nig and
                  Martin Hoffmann},
  title        = {Privacy-preserving People Detection Enabled by Solid State LiDAR},
  booktitle    = {16th Int. Conference on Intelligent Environments (IE)},
  publisher    = {{IEEE}},
  year         = {2020},
}

@inproceedings{DBLP:conf/iccv/KirillovMRMRGXW23,
  author       = {Alexander Kirillov and
                  Eric Mintun and
                  Nikhila Ravi and
                  Hanzi Mao and
                  Chlo{\'{e}} Rolland and
                  Laura Gustafson and
                  Tete Xiao and
                  Spencer Whitehead and
                  Alexander C. Berg and
                  Wan{-}Yen Lo and
                  Piotr Doll{\'{a}}r and
                  Ross B. Girshick},
  title        = {Segment Anything},
  booktitle    = {{IEEE/CVF} International Conference on Computer Vision (ICCV)},
  pages        = {3992--4003},
  year         = {2023},
}

@inproceedings{DBLP:conf/cvpr/Li0XL0NS23,
  author       = {Feng Li and
                  Hao Zhang and
                  Huaizhe Xu and
                  Shilong Liu and
                  Lei Zhang and
                  Lionel M. Ni and
                  Heung{-}Yeung Shum},
  title        = {Mask {DINO:} Towards {A} Unified Transformer-based Framework for Object
                  Detection and Segmentation},
  booktitle    = {{IEEE/CVF} Conf. on Computer Vision and Pattern Recognition (CVPR)},
  pages        = {3041--3050},
  year         = {2023},
}

@inproceedings{DBLP:conf/iccv/LiuL00W0LG21,
  author       = {Ze Liu and
                  Yutong Lin and
                  Yue Cao and
                  Han Hu and
                  Yixuan Wei and
                  Zheng Zhang and
                  Stephen Lin and
                  Baining Guo},
  title        = {Swin {Transformer}: Hierarchical Vision Transformer using Shifted Windows},
  booktitle    = {{IEEE/CVF} International Conference on Computer Vision (ICCV)},
  pages        = {9992--10002},
  year         = {2021},
}

@inproceedings{DBLP:conf/icra/DroeschelSHB11,
  author       = {David Droeschel and
                  J{\"{o}}rg St{\"{u}}ckler and
                  Dirk Holz and
                  Sven Behnke},
  title        = {Towards joint attention for a domestic service robot - person awareness
                  and gesture recognition using Time-of-Flight cameras},
  booktitle    = {{IEEE} Int. Conference on Robotics and Automation (ICRA)},
  pages        = {1205--1210},
  year         = {2011},
}

@inproceedings{DBLP:conf/cvpr/ViolaJ01,
  author       = {Paul A. Viola and
                  Michael J. Jones},
  title        = {Rapid Object Detection using a Boosted Cascade of Simple Features},
  booktitle    = {{IEEE} Conference on Computer Vision and Pattern
                  Recognition (CVPR)},
  pages        = {511--518},
  year         = {2001},
}

@inproceedings{DBLP:conf/cvpr/DalalT05,
  author       = {Navneet Dalal and
                  Bill Triggs},
  title        = {Histograms of Oriented Gradients for Human Detection},
  booktitle    = {{IEEE} Computer Society Conference on Computer Vision and Pattern
                  Recognition (CVPR)},
  pages        = {886--893},
  year         = {2005},
}

@article{DBLP:journals/pami/RenHG017,
  author       = {Shaoqing Ren and
                  Kaiming He and
                  Ross B. Girshick and
                  Jian Sun},
  title        = {Faster {R-CNN:} Towards Real-Time Object Detection with Region Proposal
                  Networks},
  journal      = {{IEEE} Transactions on Pattern Analysis and Machine Intelligence (PAMI)},
  volume       = {39},
  number       = {6},
  pages        = {1137--1149},
  year         = {2017},
  doi          = {10.1109/TPAMI.2016.2577031},
  bibsource    = {dblp computer science bibliography, https://dblp.org}
}

@inproceedings{DBLP:conf/cvpr/RedmonDGF16,
  author       = {Joseph Redmon and
                  Santosh Kumar Divvala and
                  Ross B. Girshick and
                  Ali Farhadi},
  title        = {You Only Look Once: Unified, Real-Time Object Detection},
  booktitle    = {{IEEE} Conference on Computer Vision and Pattern Recognition (CVPR)},
  year         = {2016},
}

@inproceedings{DBLP:conf/ivs/KidonoMWNM11,
  author       = {Kiyosumi Kidono and
                  Takeo Miyasaka and
                  Akihiro Watanabe and
                  Takashi Naito and
                  Jun Miura},
  title        = {Pedestrian recognition using high-definition {LIDAR}},
  booktitle    = {{IEEE} Intelligent Vehicles Symposium (IV)},
  pages        = {405--410},
  year         = {2011},
}

@inproceedings{WagnerFHB:ESANN2016,
  author       = {J{\"{o}}rg Wagner and
                  Volker Fischer and
                  Michael Herman and
                  Sven Behnke},
  title        = {Multispectral Pedestrian Detection using Deep Fusion Convolutional
                  Neural Networks},
  booktitle    = {24th European Symposium on Artificial Neural Networks (ESANN)},
  year         = {2016},
}

@inproceedings{DBLP:conf/fgr/ClapesJME20,
  author       = {Albert Clap{\'{e}}s and
                  J{\'{u}}lio C. S. Jacques J{\'{u}}nior and
                  Carla Morral and
                  Sergio Escalera},
  title        = {{ChaLearn} {LAP} 2020 Challenge on Identity-preserved Human Detection:
                  Dataset and Results},
  booktitle    = {15th {IEEE} International Conference on Automatic Face and Gesture
                  Recognition (FG)},
  pages        = {801--808},
  year         = {2020},
}

@inproceedings{DBLP:conf/iccv/Feichtenhofer0M19,
  author       = {Christoph Feichtenhofer and
                  Haoqi Fan and
                  Jitendra Malik and
                  Kaiming He},
  title        = {{SlowFast} Networks for Video Recognition},
  booktitle    = {{IEEE/CVF} International Conference on Computer Vision (ICCV)},
  pages        = {6201--6210},
  year         = {2019},
}

@article{zhang:PAMI2023visibleIR,
  title={Visible and infrared image fusion using deep learning},
  author={Zhang, Xingchen and Demiris, Yiannis},
  journal={IEEE Transactions on Pattern Analysis and Machine Intelligence (PAMI)},
  volume={45},
  number={8},
  pages={10535--10554},
  year={2023},
}

@article{tian:TMM2024RGB-Thermal,
  title={Cross-Modality Proposal-guided Feature Mining for Unregistered {RGB}-Thermal Pedestrian Detection},
  author={Tian, Chao and Zhou, Zikun and Huang, Yuqing and Li, Gaojun and He, Zhenyu},
  journal={IEEE Transactions on Multimedia (TMM)},
  year={2024},
}

@article{beyer2018deep,
  title={Deep person detection in two-dimensional range data},
  author={Beyer, Lucas and Hermans, Alexander and Linder, Timm and Arras, Kai O and Leibe, Bastian},
  journal={IEEE Robotics and Automation Letters (RA-L)},
  volume={3},
  number={3},
  pages={2726--2733},
  year={2018},
}

@inproceedings{jia:IROS2020dr,
  title={{DR-SPAAM}: A spatial-attention and auto-regressive model for person detection in {2D} range data},
  author={Jia, Dan and Hermans, Alexander and Leibe, Bastian},
  booktitle={IEEE/RSJ International Conference on Intelligent Robots and Systems (IROS)},
  pages={10270--10277},
  year={2020},
}

@article{yang2024li2former,
  title={{Li2Former}: Omni-Dimension Aggregation Transformer for Person Detection in {2-D} Range Data},
  author={Yang, Haodong and Yang, Yadong and Yao, Chenpeng and Liu, Chengju and Chen, Qijun},
  journal={IEEE Tr. on Instrumentation and Measurement (TIM)},
  year={2024},
  publisher={IEEE}
}

@inproceedings{jafari2014real,
  title={Real-time {RGB-D} based people detection and tracking for mobile robots and head-worn cameras},
  author={Jafari, Omid Hosseini and Mitzel, Dennis and Leibe, Bastian},
  booktitle={2014 IEEE International Conference on Robotics and Automation (ICRA)},
  pages={5636--5643},
  year={2014},
}

@article{xu2023onboard,
  title={Onboard dynamic-object detection and tracking for autonomous robot navigation with {RGB-D} camera},
  author={Xu, Zhefan and Zhan, Xiaoyang and Xiu, Yumeng and Suzuki, Christopher and Shimada, Kenji},
  journal={IEEE Robotics and Automation Letters (RA-L)},
  volume={9},
  number={1},
  pages={651--658},
  year={2023},
}

@article{yan2020online,
  title={Online learning for {3D} {LiDAR}-based human detection: experimental analysis of point cloud clustering and classification methods},
  author={Yan, Zhi and Duckett, Tom and Bellotto, Nicola},
  journal={Autonomous Robots},
  volume={44},
  number={2},
  pages={147--164},
  year={2020},
  publisher={Springer}
}

@inproceedings{LiuKCCZPCL23,
  author       = {Youquan Liu and
                  Lingdong Kong and
                  Jun Cen and
                  Runnan Chen and
                  Wenwei Zhang and
                  Liang Pan and
                  Kai Chen and
                  Ziwei Liu},
  title        = {Segment Any Point Cloud Sequences by Distilling Vision Foundation
                  Models},
  booktitle    = {Advances in Neural Information Processing Systems 36 (NeurIPS)},
  year         = {2023},
}

@article{hong:PAMI2024unified,
  title={Unified {3D} and {4D} panoptic segmentation via dynamic shifting networks},
  author={Hong, Fangzhou and Kong, Lingdong and Zhou, Hui and Zhu, Xinge and Li, Hongsheng and Liu, Ziwei},
  journal={IEEE Tr. on Pattern Analysis and Machine Intelligence (PAMI)},
  year={2024},
}

@inproceedings{DBLP:conf/iros/MemmesheimerTP20,
  author       = {Raphael Memmesheimer and
                  Nick Theisen and
                  Dietrich Paulus},
  title        = {Gimme {Signals}: Discriminative signal encoding for multimodal activity
                  recognition},
  booktitle    = {{IEEE/RSJ} International Conference on Intelligent Robots and Systems (IROS)},
  pages        = {10394--10401},
  year         = {2020},
}

@inproceedings{DBLP:conf/icpr/MemmesheimerTP20,
  author       = {Raphael Memmesheimer and
                  Nick Theisen and
                  Dietrich Paulus},
  title        = {{SL-DML}: Signal Level Deep Metric Learning for Multimodal One-Shot
                  Action Recognition},
  booktitle    = {25th International Conference on Pattern Recognition (ICPR)},
  pages        = {4573--4580},
  publisher    = {{IEEE}},
  year         = {2020},
}

@inproceedings{DBLP:conf/ccs/HitajAP17,
  author       = {Briland Hitaj and
                  Giuseppe Ateniese and
                  Fernando P{\'{e}}rez{-}Cruz},
  title        = {Deep Models Under the {GAN:} Information Leakage from Collaborative
                  Deep Learning},
  booktitle    = {{ACM} {SIGSAC} Conference on Computer and
                  Communications Security (CCS)},
  pages        = {603--618},
  year         = {2017},
}

@article{li2023multi,
  title={Multi-step jailbreaking privacy attacks on {ChatGPT}},
  author={Li, Haoran and Guo, Dadi and Fan, Wei and Xu, Mingshi and Song, Yangqiu},
  journal={preprint arXiv:2304.05197},
  year={2023}
}

@inproceedings{DBLP:conf/iccv/DaiQXLZHW17,
  author       = {Jifeng Dai and
                  Haozhi Qi and
                  Yuwen Xiong and
                  Yi Li and
                  Guodong Zhang and
                  Han Hu and
                  Yichen Wei},
  title        = {Deformable Convolutional Networks},
  booktitle    = {{IEEE} International Conference on Computer Vision (ICCV)},
  pages        = {764--773},
  year         = {2017},
}

@inproceedings{DBLP:conf/iccv/BehleyGMQBSG19,
  author       = {Jens Behley and
                  Martin Garbade and
                  Andres Milioto and
                  Jan Quenzel and
                  Sven Behnke and
                  Cyrill Stachniss and
                  J{\"{u}}rgen Gall},
  title        = {SemanticKITTI: {A} Dataset for Semantic Scene Understanding of {LiDAR}
                  Sequences},
  booktitle    = {{IEEE/CVF} International Conference on Computer Vision (ICCV)},
  pages        = {9296--9306},
  year         = {2019},
}

@article{DBLP:journals/cacm/ShottonSKFFBCM13,
  author       = {Jamie Shotton and
                  Toby Sharp and
                  Alex Kipman and
                  Andrew W. Fitzgibbon and
                  Mark Finocchio and
                  Andrew Blake and
                  Mat Cook and
                  Richard Moore},
  title        = {Real-time human pose recognition in parts from single depth images},
  journal      = {Communications of the {ACM}},
  volume       = {56},
  number       = {1},
  pages        = {116--124},
  year         = {2013},
}

@article{DBLP:journals/ras/WangWLMY17,
  author       = {Heng Wang and
                  Bin Wang and
                  Bingbing Liu and
                  Xiaoli Meng and
                  Guang{-}Hong Yang},
  title        = {Pedestrian recognition and tracking using {3D} {LiDAR} for autonomous
                  vehicle},
  journal      = {Robotics and Autonomous Systems (RAS)},
  volume       = {88},
  pages        = {71--78},
  year         = {2017},
}

@inproceedings{cortinhal2020salsanext,
  title={{SalsaNext}: Fast, Uncertainty-Aware Semantic Segmentation of {LiDAR} Point Clouds},
  author={Cortinhal, Tiago and Tzelepis, George and Erdal Aksoy, Eren},
  booktitle={15th International Symposium on Advances in Visual Computing (ISVC)},
  pages={207--222},
  year={2020},
}

\end{document}